\documentclass{llncs}

\usepackage[a4paper,includeheadfoot,top=1.65in,bottom=1.65in,left=1.73in,hcentering]{geometry}
\usepackage[pdftex]{graphicx}
\usepackage[sectionbib]{natbib}
\usepackage{caption} 
\usepackage[utf8]{inputenc}
\usepackage{t1enc}
\usepackage{tikz-dependency}
\usepackage{qtree}
\usepackage{float}
\usepackage[title]{appendix}

\usepackage{booktabs}
\usepackage{array}
\usepackage{multirow}

\newcommand{\rescolwidth}{0.05}

\usepackage{subfig}
\restylefloat{figure}
\usepackage{tikz-qtree}
\usepackage[english]{babel}
\renewcommand\bibname{References}

\newif{\ifhidecomments}
\hidecommentsfalse

\ifhidecomments
    \newcommand{\bence}[1]{}
    \newcommand{\attila}[1]{}
    \newcommand{\judit}[1]{}
\else
    \newcommand{\bence}[1]{\textcolor{red}{[#1 ({\bf Bence})]}}
    \newcommand{\attila}[1]{\textcolor{blue}{[#1 ({\bf Attila})]}} 
    \newcommand{\judit}[1]{\textcolor{orange}{[#1 ({\bf Judit})]}} 
\fi

\begin{document}

\title{Automatic punctuation restoration with BERT models}
	\author{Attila Nagy\inst{1}, Bence Bial\inst{1}, Judit \'Acs\inst{1}\\
 \institute{
 $^1$Department of Automation and Applied Informatics\\
Budapest University of Technology and Economics}
}

\maketitle

\begin{abstract}
We present an approach for automatic punctuation restoration with BERT models for English and Hungarian. For English, we conduct our experiments on Ted Talks, a commonly used benchmark for punctuation restoration, while for Hungarian we evaluate our models on the Szeged Treebank dataset. Our best models achieve a macro-averaged $F_1$-score of 79.8 in English and 82.2 in Hungarian. Our code is publicly available\footnote{https://github.com/attilanagy234/neural-punctuator}.

\end{abstract}

\section{Introduction}
Automatic Speech Recognition (ASR) systems typically output unsegmented transcripts without punctuation.
Restoring punctuations is an important step in processing transcribed speech.
\cite{tundik2018user} showed that the absence of punctuations in transcripts affect readability as much as a
significant word error rate. Downstream tasks such as neural machine translation \citep{vandeghinste2018comparison}, sentiment analysis \citep{cureg2019sentiment} and information extraction \citep{makhoul2005effects} also benefit from having clausal boundaries.
In this paper we present models for automatic punctuation restoration for English and Hungarian.
Our work is based on a state-of-the-art model proposed by \citep{courtland2020efficient}, which uses pretrained contextualized language models \citep{devlin2018bert}.

Our contributions are twofold.
First, we present the implementation of an automatic punctuation model based on a state-of-the-art model \citep{courtland2020efficient} and evaluate it on an English benchmark dataset.
Second, using the same architecture we propose an automatic punctuator for Hungarian trained on the Szeged Treebank \citep{csendes2005szeged}.
To the best of our knowledge our work is the first punctuation restoration attempt that uses BERT on Hungarian data.

\section{Related Work}
Systems that are most efficient at restoring punctuations usually exploit both prosodic and lexical features with hybrid models \citep{szaszak2019leveraging, garg2018analysis, zelasko2018punctuation}.
Up until the appearance of BERT-like models, lexical features were primarily processed by recurrent neural networks \citep{vandeghinste2018comparison, tundik2017bilingual, kim2019deep, tilk2016bidirectional, salloum2017deep}, while more recent approaches use the transformer \citep{vaswani2017attention} architecture \citep{chen2020controllable, nguyen2019fast, cai2019question}.
The current state-of-the-art method by \cite{courtland2020efficient} is a pretrained BERT, which aggregates multiple predictions for the same token, resulting in higher accuracy and significant parallelism.

\section{Methodology}
We train models for Hungarian and English. For English we rely on the widely used IWSLT 2012 Ted Talks dataset \citep{federico2012overview} benchmark dataset.
Due to the lack of such datasets for Hungarian, we generate it from the Szeged Treebank \citep{csendes2005szeged}.
We preprocess the Szeged Treebank such that it structures similarly to the output of an ASR system. Then with the presented methods we attempt to reconstruct the original and punctuated gold standard corpus.

\subsection{Problem formulation}
We formulate the problem of punctuation restoration as a sequence labeling task with four target classes: \emph{EMPTY}, \emph{COMMA}, \emph{PERIOD}, and \emph{QUESTION}. We do not include other punctuation marks as their frequency is very low in both datasets.
For this reason, we apply a conversion in cases where it is semantically reasonable: we convert exclamation marks and semicolons to periods and colons and quotation marks to commas.
We remove double and intra-word hyphens, however, if they are encapsulated between white spaces, we convert them to commas. Other punctuation marks are disregarded during our experiments. As tokenizers occasionally split words to multiple tokens, we apply masking on tokens, which do not mark a word ending. These preprocessing steps and the corresponding output labels are shown in Table \ref{table:example-input-sentence}.

\begin{table*}[ht]
\centering
\begin{tabular}{@{}llccccc@{}}
\toprule
Original & \multicolumn{6}{l}{Tyranosaurus asked: kill me?} \\ \midrule
\multicolumn{1}{l|}{Preprocessed} & \multicolumn{6}{l}{tyranosaurus asked, kill me?} \\
\multicolumn{1}{l|}{Tokenized} & \multicolumn{1}{c}{ty} & \#\#rano & \#\#saurus & asked & kill & me \\
\multicolumn{1}{l|}{Output} & \multicolumn{1}{c}{-} & - & EMP & COM & EMP & Q \\ \midrule
\multicolumn{1}{l|}{Original} & \multicolumn{6}{l}{Not enough, -- said the co-pilot -- ...} \\
\midrule
\multicolumn{1}{l|}{Preprocessed} & \multicolumn{6}{l}{not enough, said the co pilot,} \\
\multicolumn{1}{l|}{Tokenized} & \multicolumn{1}{c}{not} & enough & said & the & co & pilot \\
\multicolumn{1}{l|}{Output} & \multicolumn{1}{c}{EMP} & COM & EMP & EMP & EMP & COM \\ \bottomrule
\end{tabular}
\caption{An example input sentence and the following processing steps in our setup.}
\label{table:example-input-sentence}
\end{table*}

\subsection{Datasets}

\subsubsection{IWSLT 2012 Ted Talks dataset}
We use the IWSLT 2012 Ted Talks dataset \citep{federico2012overview} for English.
IWSLT is a common benchmark for automatic punctuation.
It contains 1066 unique transcripts of Ted talks with a total number of 2.46M words in the corpus.
We lowercase the data and we convert consecutive spaces into single spaces.
We also remove spaces before commas.
We use the original train, validation and test sets from the IWSLT 2012 competition.
The overall data distributions of the IWSLT Ted Talk dataset is summarized in Table~\ref{table:ted-talk-summary}.

\begin{table*}[ht]
\centering
\begin{tabular}{lrrr}
\toprule
 & Train & Validation & Test \\
\midrule
PERIOD & 139,619 & 909 & 1,100 \\
COMMA & 188,165 & 1,225 & 1,120 \\
QUESTION & 10,215 & 71 & 46 \\
EMPTY & 2,001,462 & 15,141 & 16,208 \\
\bottomrule
\end{tabular}
\caption{Label distributions of the IWSLT Ted talk dataset.}
\label{table:ted-talk-summary}
\end{table*}

\subsubsection{Szeged Treebank}
We use the Szeged Treebank dataset \citep{csendes2005szeged} for Hungarian.
This dataset is the largest gold standard treebank in Hungarian.
It covers a wide variety of domains such as fiction, news articles, and legal text.
As these subcorpora have very different distributions in terms of punctuations, we merge them and shuffle the sentences.
We then split the dataset into train, validation and test sets.
This introduces a bias in the prediction of periods as it is easier for the model to correctly predict sentence boundaries by recognizing context change between adjacent sentences but it also provides a more-balanced distribution of punctuation classes across the train, validation and test sets.
The label distribution is listed in Table~\ref{table:szeged-treebank-summary}.

\begin{table*}[ht]
\centering
\begin{tabular}{lrrr}
\toprule
 & Train & Validation & Test \\
\midrule
PERIOD & 81,168 & 9,218 & 3,370 \\
COMMA & 120,027 & 13,781 & 4,885 \\
QUESTION & 1,808 & 198 & 75 \\
EMPTY & 885,451 & 101,637 & 36,095 \\
\bottomrule
\end{tabular}
\caption{Overall data distributions of the Szeged Treebank dataset.}
\label{table:szeged-treebank-summary}
\end{table*}

\subsection{Architecture}
\begin{figure}
    \centering
    \includegraphics[width=0.75\textwidth]{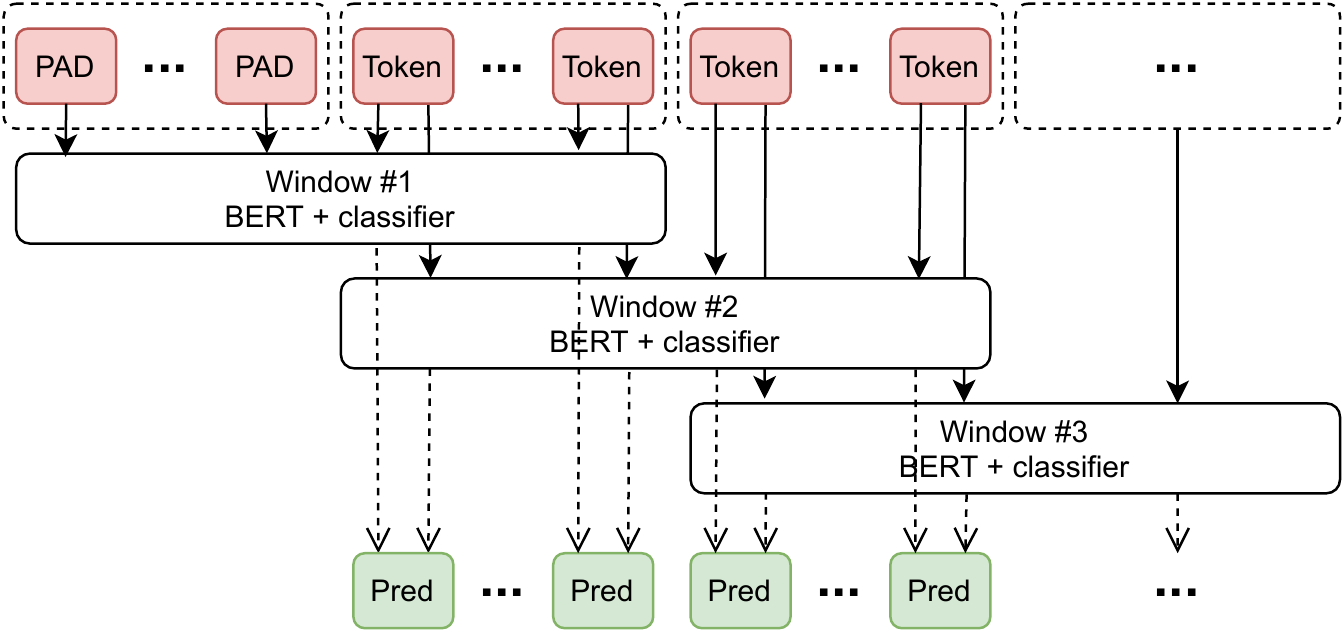}
    \caption{The process of generating multiple predictions for a token. Although BERT always receives sequences of 512, we sample consecutive sequences from the corpora such that they overlap, thus resulting in multiple predictions for the same token. The extent of the overlap and therefore the number of predictions for a token depend on the offset between the windows. Note that padding is necessary in the beginning to ensure that all tokens have the same amount of predictions.}
    \label{fig:pred-box}
\end{figure}

\begin{figure}
    \centering
     \includegraphics[width=0.4\textwidth]{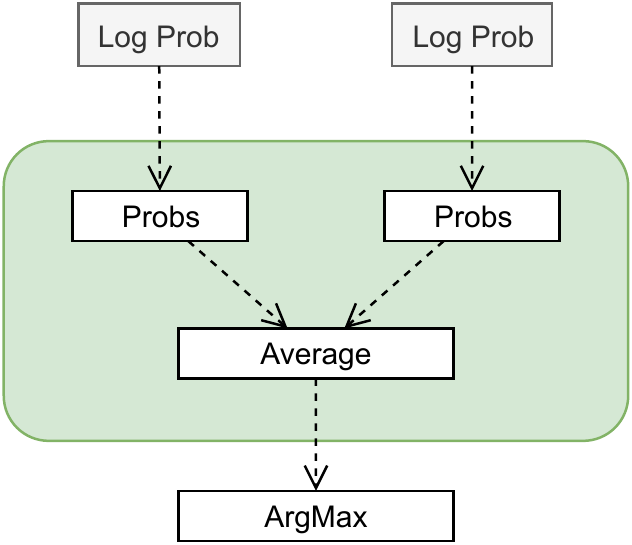}
    \caption{The final prediction is computed by first aggregating all punctuation probability distributions for each token by taking their class-wise averages and then selecting the highest probability.}
    \label{fig:window}
\end{figure}

\begin{figure}
    \centering
     \includegraphics[width=0.9\textwidth]{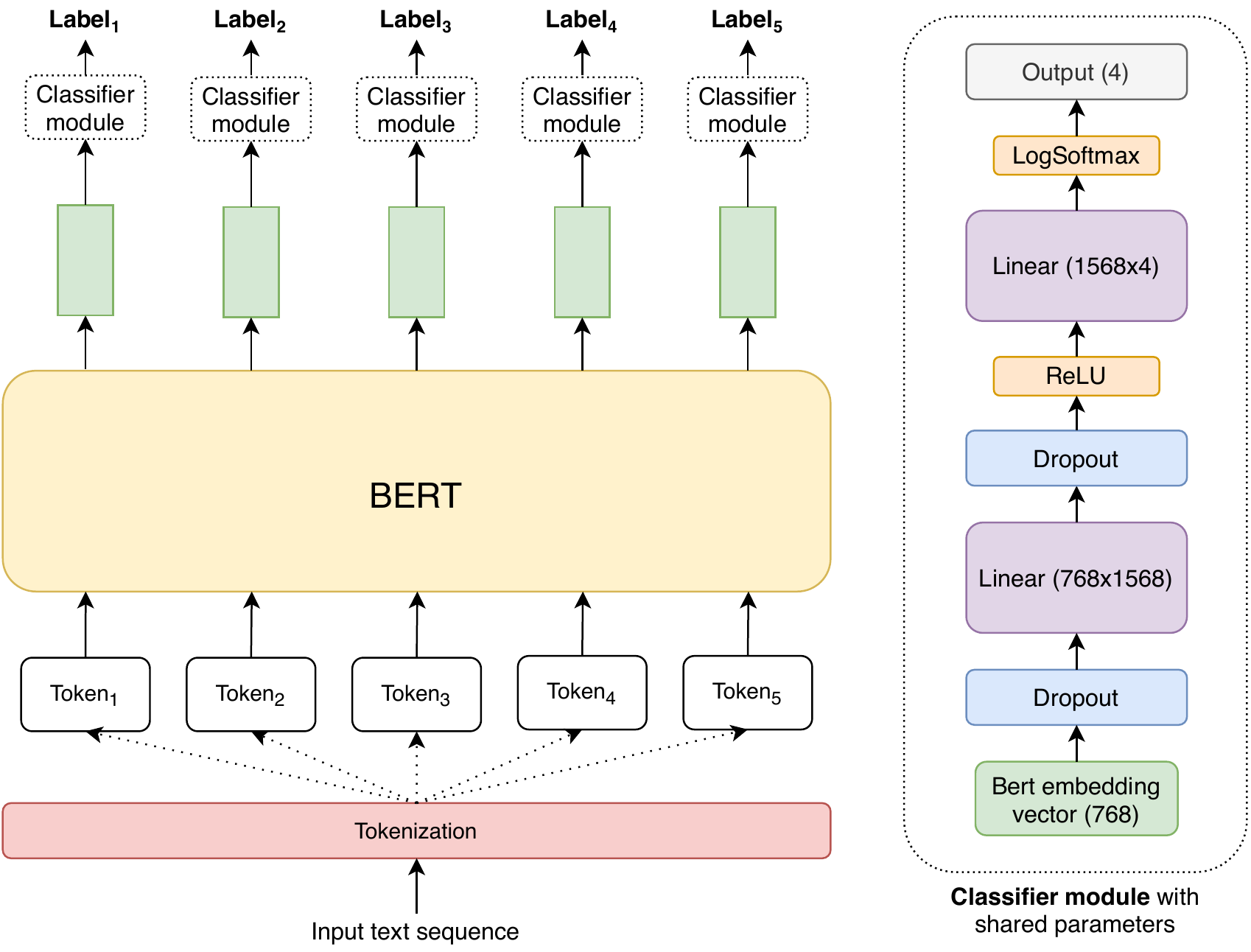}
    \caption{The complete architecture used for punctuation restoration.}
    \label{fig:punctuator-arch}
\end{figure}

Our model is illustrated in Figure~\ref{fig:punctuator-arch}.
We base our model on pretrained BERT models.
BERT is a contextual language model with multiple transformer layers and hundreds of millions trainable parameters trained on a massive English corpora with the masked language modeling objective.
Several variations of pretrained weights were released.
We use BERT-base cased and uncased for English as well as Albert \citep{lan2019albert}, a smaller version of BERT.
BERT also has a multilingual version, \emph{mBERT} that supports Hungarian along with 100 other languages.
We use mBERT and the recently released Hungarian-only BERT, \emph{huBERT} \citep{Nemeskey:2020} for Hungarian.
These models all apply wordpiece tokenization with their own predefined WordPiece vocabulary.
They then generate continuous representations for every wordpiece.
Our model adds a two-layer multilayer perceptron on top of these representation with 1568 hidden dimension, ReLU activation and an output layer, and finally a softmax layer that produces a distribution over the labels.
We also apply dropout with a probability of 0.2 before and after the first linear layer. Similarly to \cite{courtland2020efficient}, we apply a sliding window over the input data, generate multiple predictions for each token and then aggregate the probabilities for each position by taking the label-wise mean and thus output the most probable label. The process is illustrated in Figure \ref{fig:pred-box} and \ref{fig:window}.

\subsection{Training}
We train all models with identical hyperparameters.
We perform gradient descent using the AdamW optimizer \citep{loshchilov2017decoupled} with the learning rate set to $3 * 10^{-5}$ for BERT and $10^{-4}$ for the classifier on top.
We apply gradient clipping of 1.5 and a learning rate warm up of 300 steps using a linear scheduler.
We select negative log likelihood as the loss function.
The tokenizer modules often split single words to multiple subwords. For this task we only need to predict punctuations after words (between white spaces).
We mask the loss function for every other subword.
It is a common practice to intermittently freeze and unfreeze the weights of the transformer model, while training the fine-tuning linear layers situated at the top of the whole architecture.
We found that it is best to have the transformer model unfrozen from the very first epoch and therefore update its parameters along with the linear layers.
We trained the models for 12 epochs with a batch size of 4 and applied early stopping based on the validation set.
We used the validation set to tune the sliding window step size, that is responsible for getting multiple predictions for a single token.
All experiments were performed using a single Nvidia GTX 1070 GPU with one epoch taking 10 minutes.
Our longest training lasted for 2 hours.

\section{Results}
All models are evaluated using macro $F_1$-score ($F$) over the 4 classes. Similarly to Courtland et al., our work is focused on the performance of punctuation marks and as \textit{EMPTY} labels constitute 85\% of all labels, we report the overall $F_1$-score without \textit{EMPTY}.
We evaluated both cased and uncased variations of BERT and generally we have found that the uncased model is better than its cased variant for this task.
This was an expected conclusion, as we lowercased the entire corpus with the purpose of eliminating bias around the prediction of periods.
For all setups, we selected the best performing models on the validation set by loss and by macro $F_1$-score and evaluated them independently on the test set. On the Ted Talks dataset, our best performing model was an uncased variation of BERT that achieved on par performance with the current state-of-the-art model \citep{courtland2020efficient}, having a slightly worse macro $F_1$-score of 79.8 (0.8 absolute and 0.9975\% relative difference) with 10 epochs of training and 64 predictions/token. All results on the Ted Talks dataset are summarized in Table \ref{table:results_table_ted}.

\begin{table*}
\small
    \centering
    \begin{tabular}{p{0.28\textwidth}|
            >{\centering}p{\rescolwidth\textwidth}
            >{\centering}p{\rescolwidth\textwidth}
            >{\centering}p{\rescolwidth\textwidth}
            |
            >{\centering}p{\rescolwidth\textwidth}
            >{\centering}p{\rescolwidth\textwidth}
            >{\centering}p{\rescolwidth\textwidth}
            |
            >{\centering}p{\rescolwidth\textwidth}
            >{\centering}p{\rescolwidth\textwidth}
            >{\centering}p{\rescolwidth\textwidth}
            |
            >{\centering}p{\rescolwidth\textwidth}
            >{\centering}p{\rescolwidth\textwidth}
            c
    }
        \toprule
        & \multicolumn{3}{c}{Comma} & \multicolumn{3}{c}{Period} & \multicolumn{3}{c}{Question} & \multicolumn{3}{c}{Overall}\\
        Models & P & R & F & P & R & F & P & R & F & P & R & F \\
        \midrule
        BERT-base \citep{courtland2020efficient} & \textbf{72.8} & 70.8 & \textbf{71.8} & 81.9 & 86.6 & \textbf{84.2} & 80.8 & \textbf{91.3} & 85.7 & \textbf{78.5} & 82.9 & \textbf{80.6} \\
        Albert-base \citep{courtland2020efficient} & 69.4 & 69.3 & 69.4 & 80.9 & 84.5 & 82.7 & 76.7 & 71.7 & 74.2 & 75.7 & 75.2 & 75.4 \\
        BERT-base-uncased (by loss) & 59.0 & 80.2 & 68 & 83.0 & 83.6 & 83.3 & 87.8 & 83.7 & 85.7 & 76.6 & 82.5 & 79.0 \\
        BERT-base-uncased (by $F_1$-score) & 58.4 & \textbf{80.7} & 67.8 & \textbf{84.2} & 83.8 & 84.0 & \textbf{84.8} & 90.7 & \textbf{87.6} & 75.8 & \textbf{85.1} & 79.8 \\
        BERT-base-cased (by loss) & 57.3 & 73.9 & 64.5 & 75.9 & 87.9 & 81.4 & 77.1 & 84.1 & 80.4 & 70.1 & 81.9 & 75.5 \\
        BERT-base-cased (by $F_1$-score) & 59.1 & 78.5 & 67.5 & 79.6 & 81.6 & 80.6 & 76.9 & 88.9 & 82.5 & 71.9 & 83.0 & 76.8 \\
        Albert-base (by loss) & 55.3 & 74.8 & 63.6 & 76.8 & \textbf{87.9} & 82.0 & 70.6 & 83.7 & 76.6 & 67.6 & 82.1 & 74.1 \\
        Albert-base (by $F_1$-score) & 56.5 & 80.3 & 66.3 & 80.7 & 80.8 & 80.8 & 80.4 & 84.1 & 82.2 & 72.5 & 81.7 & 76.4 \\
        \bottomrule
    \end{tabular}
    \caption{Precision, recall and $F_1$-score values on the Ted Talk dataset.}
    \label{table:results_table_ted}
\end{table*}

\begin{figure}%
    \centering
    \subfloat[\centering Macro $F_1$-score]{{\includegraphics[trim={1cm 0.5cm 1cm 1cm}, width=0.45\textwidth]{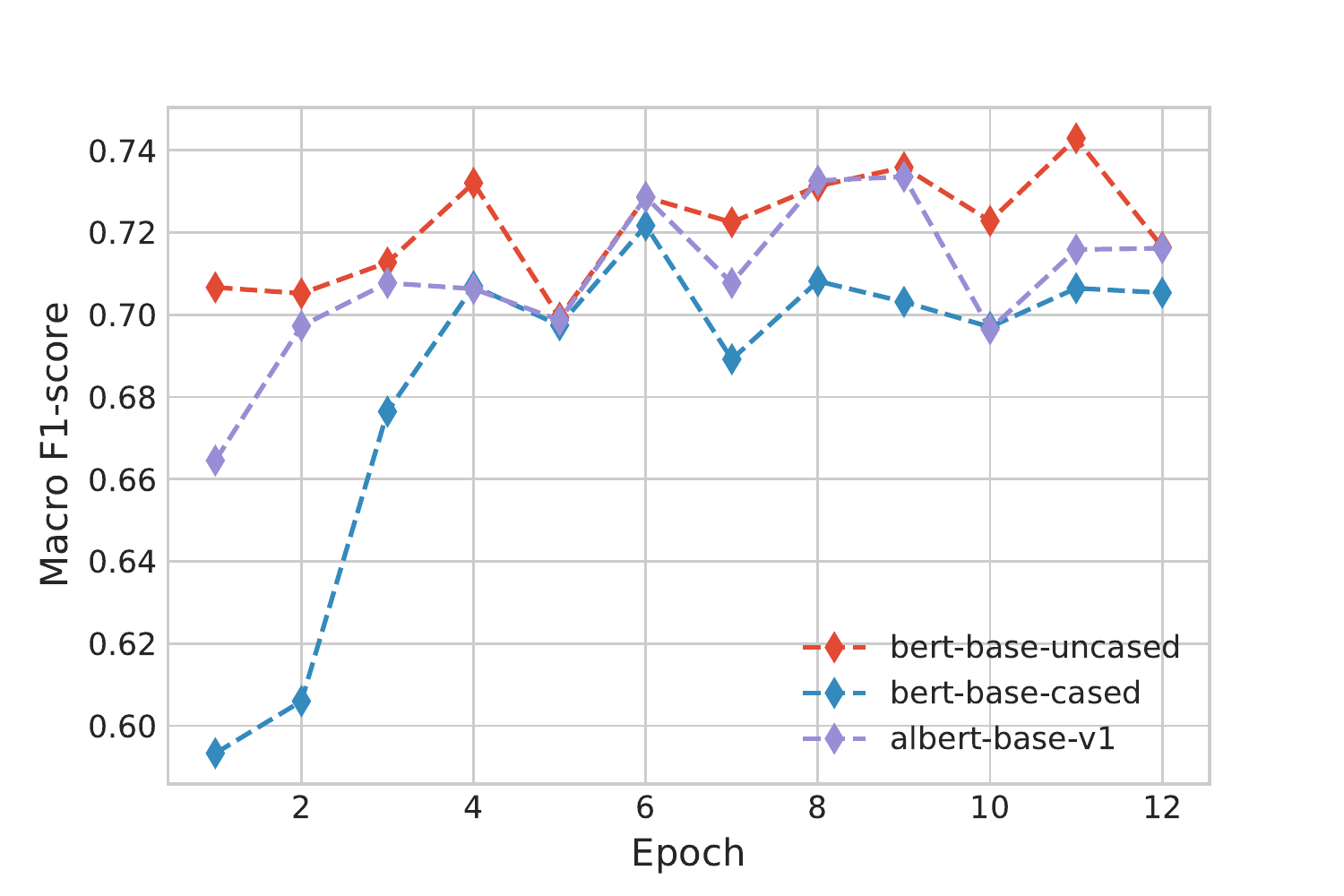} }}%
    \qquad
    \subfloat[\centering Loss]{{\includegraphics[trim={1cm 0.5cm 1cm 1cm}, width=0.45\textwidth]{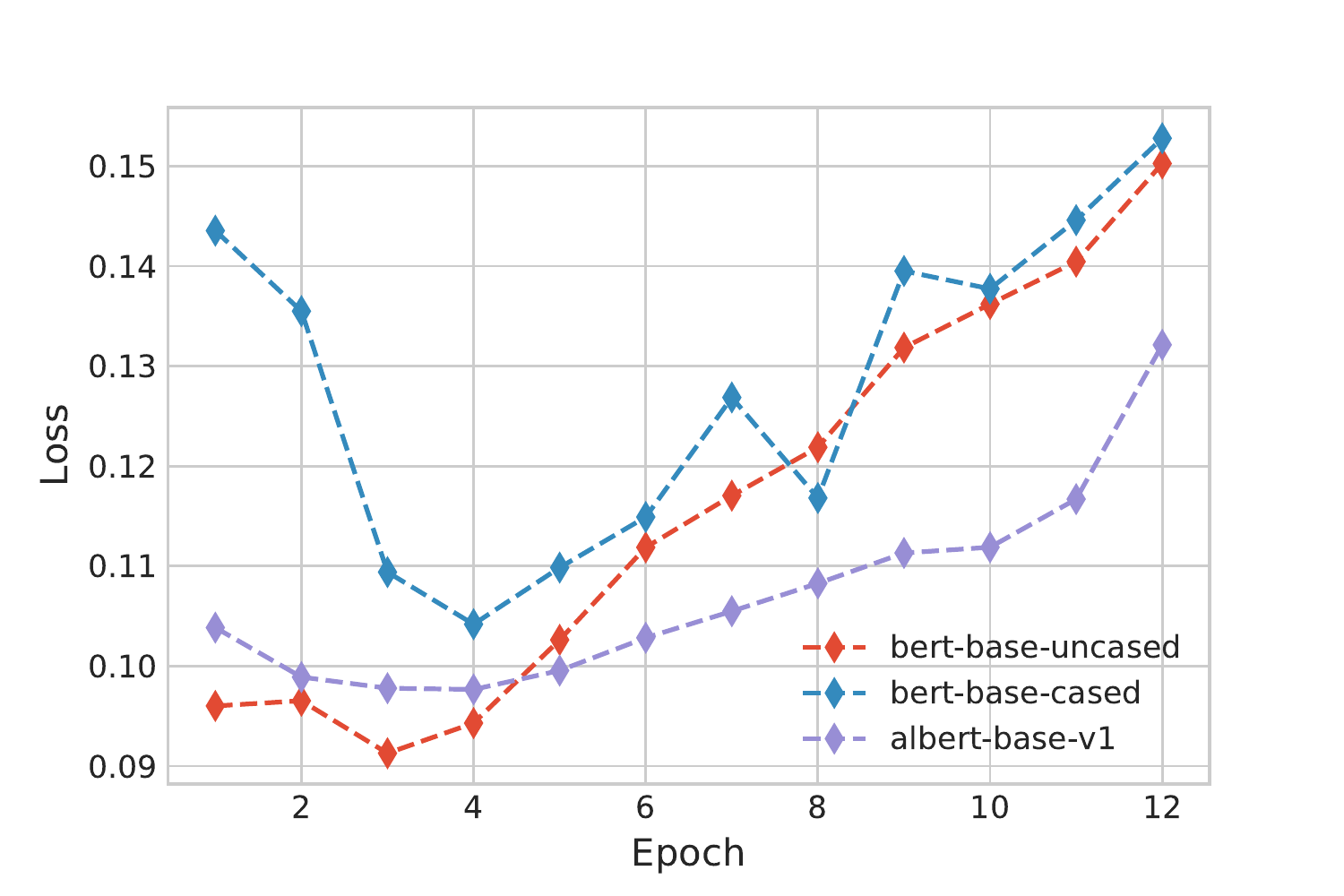} }}%
    \caption{Metrics on the validation set over epochs during training on the IWSLT Ted Talk dataset}%
    \label{fig:example}%
\end{figure}

On the Szeged Treebank dataset, we evaluate the multilangual variants of BERT and the recently released Hubert model. We find that Hubert performs significantly better (82.2 macro $F_1$-score) than the best multilangual model with an absolute and relative difference of 12.2 and 14.84\% respectively on macro $F_1$-score.
We trained the best Hubert model for 3 epochs and used 8 predictions/token. All results on the Szeged Treebank dataset are summarized in Table \ref{table:results_table_szeged}.

\begin{table*}
\small
    \centering
    \begin{tabular}{p{0.28\textwidth}|
            >{\centering}p{\rescolwidth\textwidth}
            >{\centering}p{\rescolwidth\textwidth}
            >{\centering}p{\rescolwidth\textwidth}
            |
            >{\centering}p{\rescolwidth\textwidth}
            >{\centering}p{\rescolwidth\textwidth}
            >{\centering}p{\rescolwidth\textwidth}
            |
            >{\centering}p{\rescolwidth\textwidth}
            >{\centering}p{\rescolwidth\textwidth}
            >{\centering}p{\rescolwidth\textwidth}
            |
            >{\centering}p{\rescolwidth\textwidth}
            >{\centering}p{\rescolwidth\textwidth}
            c
    }
        \toprule
        & \multicolumn{3}{c}{Comma} & \multicolumn{3}{c}{Period} & \multicolumn{3}{c}{Question} & \multicolumn{3}{c}{Overall}\\
        Models & P & R & F & P & R & F & P & R & F & P & R & F \\
        \midrule
        BERT-base-multilang-uncased (by loss) & 82.3 & 79.3 & 80.8 & 79.6 & 88.3 & 83.8 & 43.2 & 21.3 & 28.6 & 68.4 & 63.0 & 64.4 \\
        BERT-base-multilang-uncased (by $F_1$-score) & 82.9 & 79.4 & 81.1 & 80.1 & 88.4 & 84.0 & 51.4 & 24.0 & 32.7 & 71.5 & 63.9 & 66.0 \\
        BERT-base-multilang-cased (by loss) & 81.3 & 79.3 & 80.3 & 82.4 & 83.2 & 82.8 & 51.6 & 21.3 & 30.2 & 71.8 & 61.3 & 64.4 \\
        BERT-base-multilang-cased (by $F_1$-score) & 83.6 & 78.8 & 81.1 & 81.7 & 85.5 & 83.6 & 61.4 & 36.0 & 45.4 & 75.6 & 66.8 & 70.0 \\
        Hubert (by loss and $F_1$-score) & \textbf{84.4} & \textbf{87.3} & \textbf{85.8} & \textbf{89.0} & \textbf{93.1} & \textbf{91.0} & \textbf{73.5} & \textbf{66.7} & \textbf{69.9} & \textbf{82.3} & \textbf{82.4} & \textbf{82.2} \\
        \bottomrule
    \end{tabular}
    \caption{Precision, recall and $F_1$-score values on the Szeged Treebank dataset.}
    \label{table:results_table_szeged}
\end{table*}

\begin{figure}%
    \centering
    \subfloat[\centering Macro $F_1$-score]{{\includegraphics[trim={1cm 0.5cm 1cm 1cm}, width=0.45\textwidth]{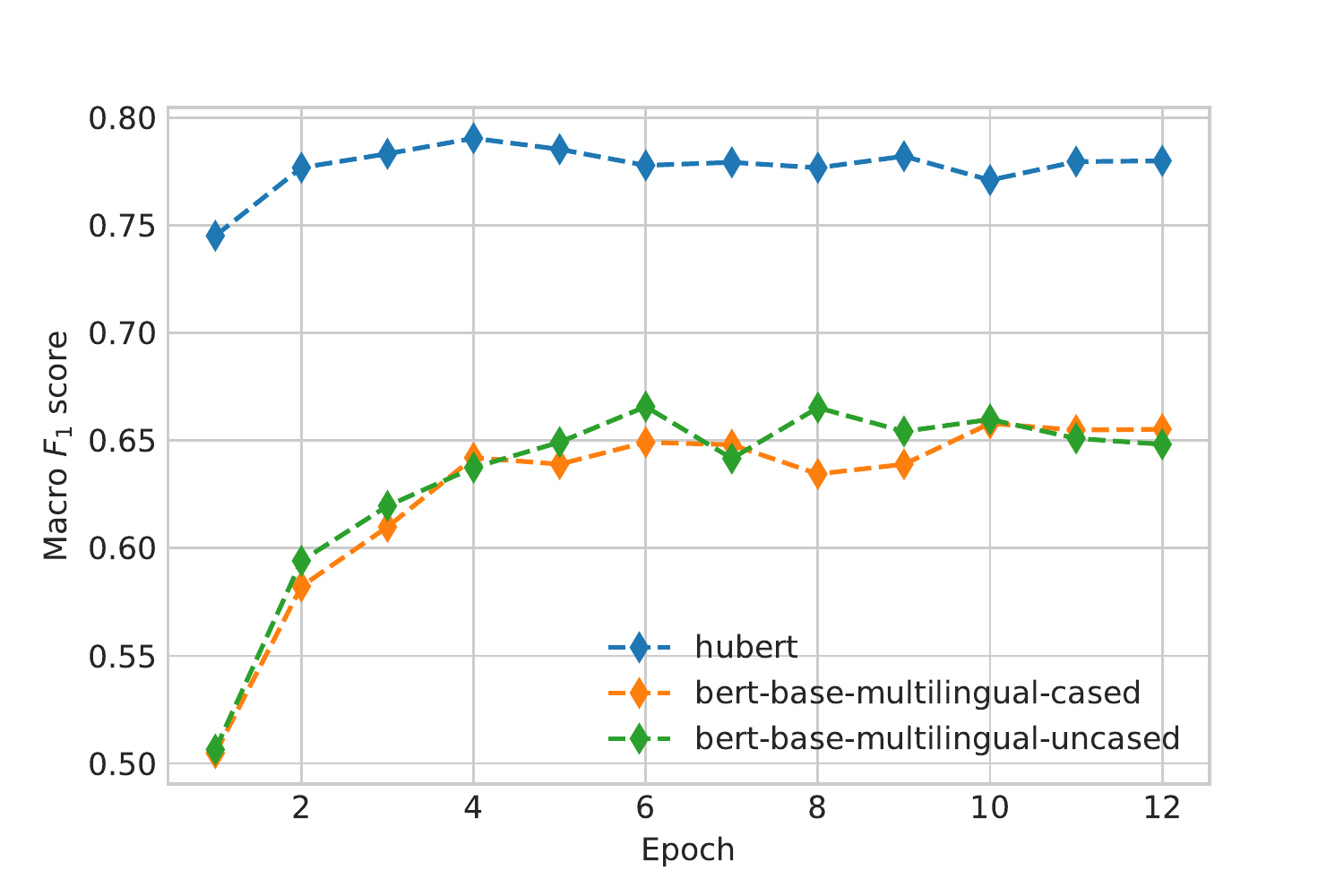} }}%
    \qquad
    \subfloat[\centering Loss]{{\includegraphics[trim={1cm 0.5cm 1cm 1cm}, width=0.45\textwidth]{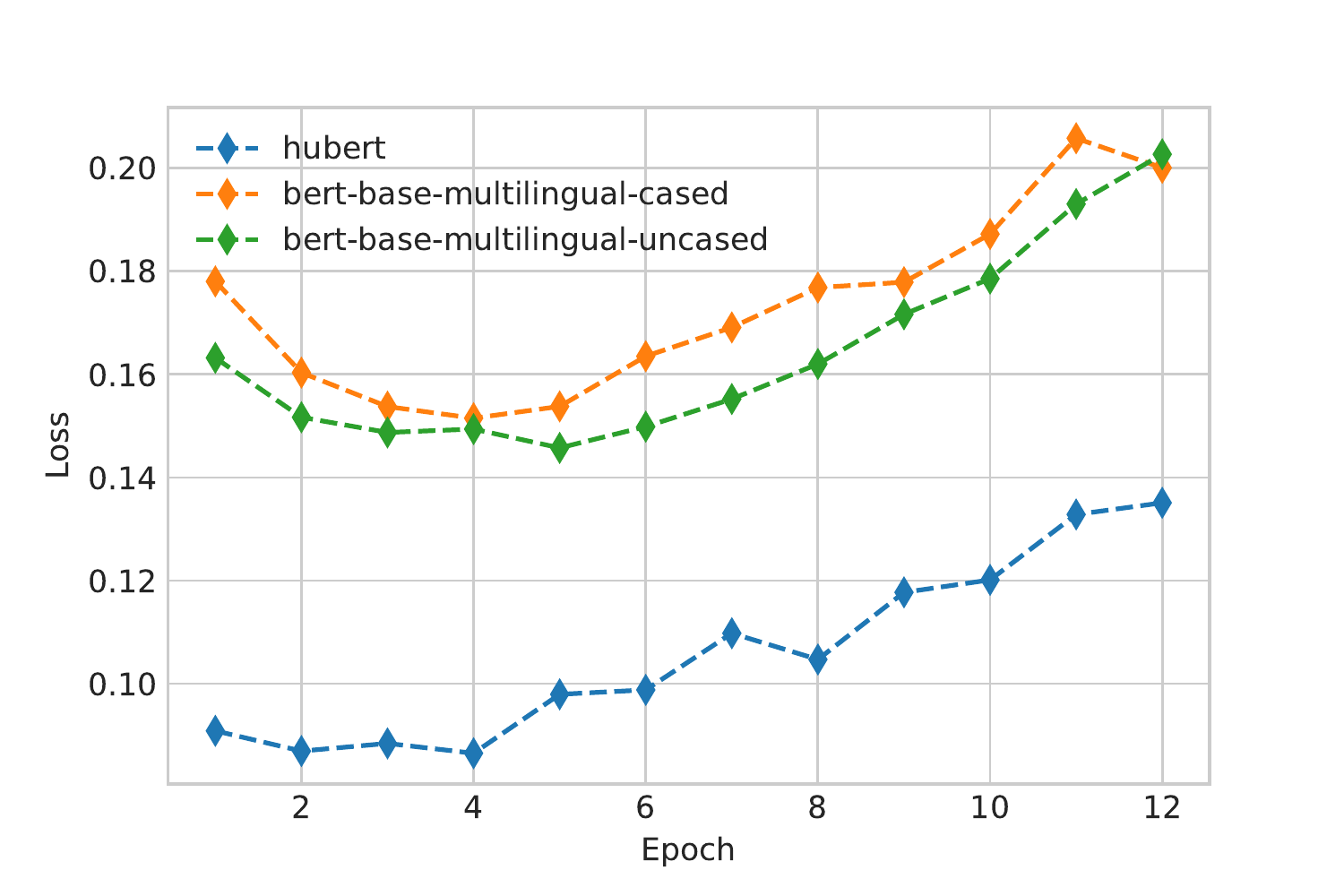} }}%
    \caption{Metrics on the validation set over epochs during training on the Szeged Treebank dataset.}%
    \label{fig:example}%
\end{figure}

\begin{figure}%
    \centering
    \subfloat[\centering BERT-base-uncased (Ted Talks) ]{{\includegraphics[trim={1cm 0.5cm 1cm 1cm}, width=0.45\textwidth]{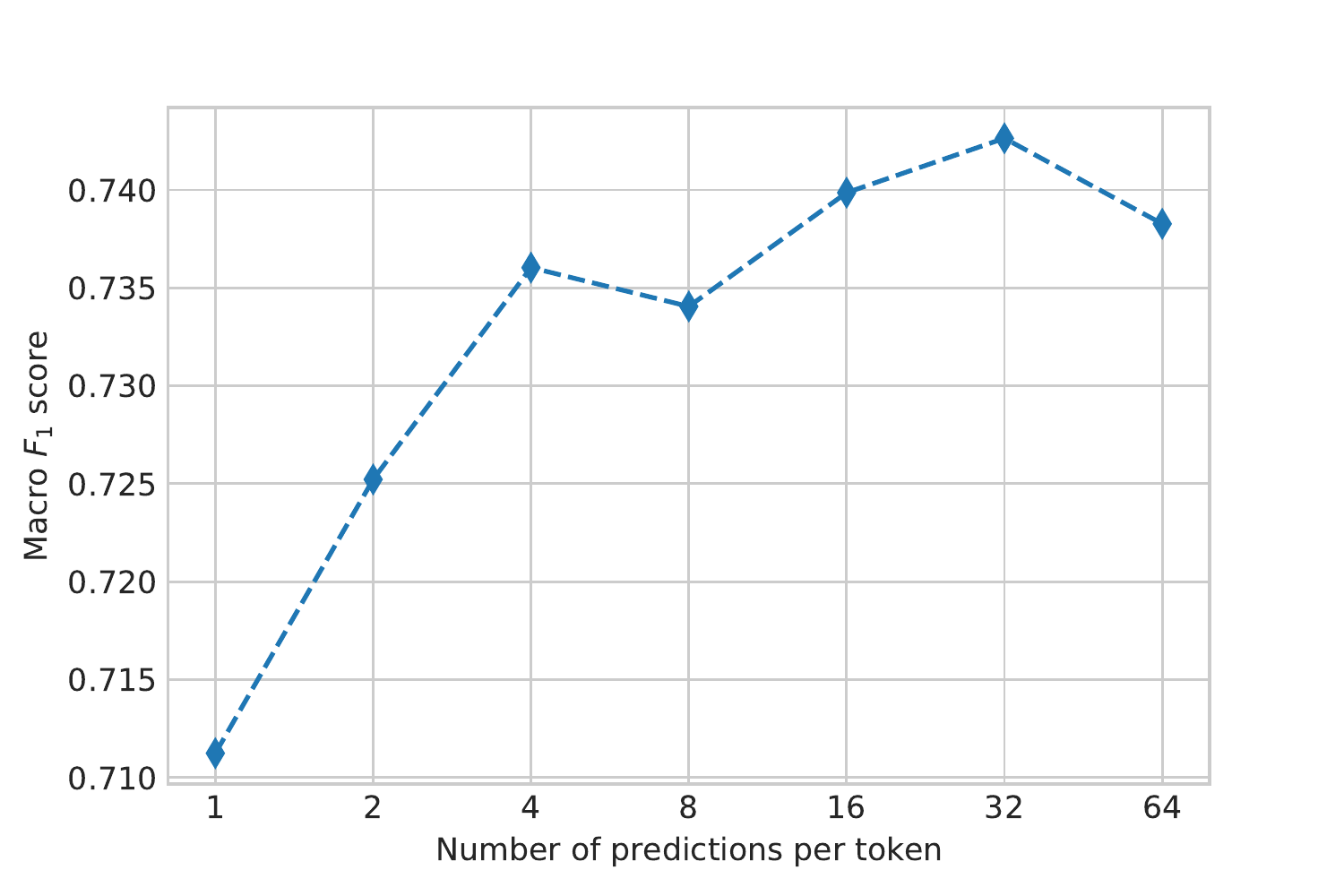} }}%
    \qquad
    \subfloat[\centering HuBERT (Szeged Treebank) ]{{\includegraphics[trim={1cm 0.5cm 1cm 1cm}, width=0.45\textwidth]{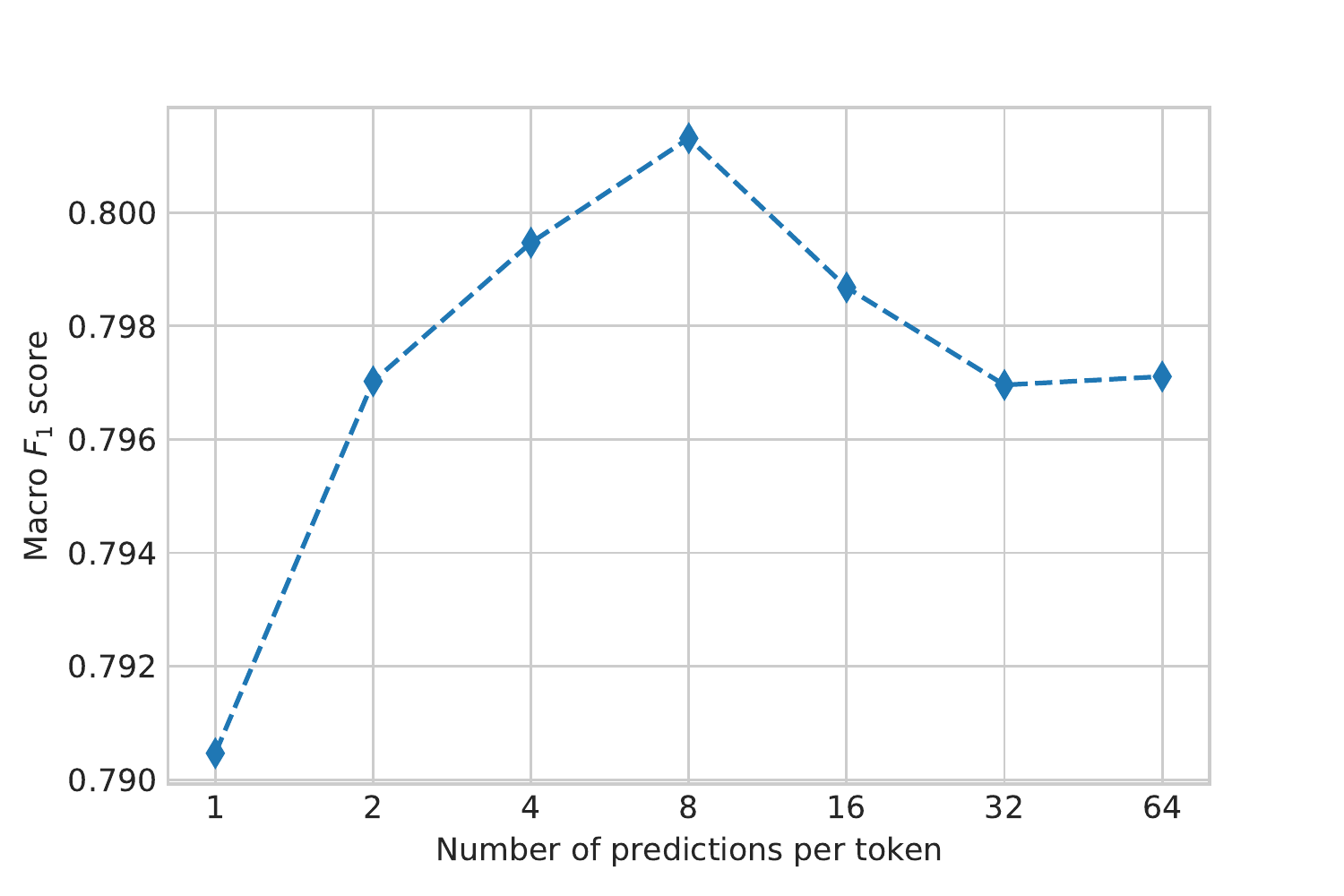} }}%
    \caption{Effect of the number of predictions per token on the overall $F_1$-score, computed on the validation dataset.}%
    \label{fig:multiple_preds}%
\end{figure}

We also examined the effect of using multiple predictions for a token. The changes we see in macro $F_1$-score on the validation set with regard to the number of predictions per token are shown in Figure~\ref{fig:multiple_preds}. 
The best models were evaluated on the test set and we found that having multiple predictions per token increased the $F_1$-score by 5\% in English and 2.4\% in Hungarian.

\section{Conclusion}
We presented an automatic punctuation restoration model based on BERT for English and Hungarian. For English we reimplemented a state-of-the-art model and evaluated it on the IWSLT Ted Talks dataset.
Our best model achieved comparable results with current state-of-the-art on the benchmark dataset.
For Hungarian we generated training data by converting the Szeged Treebank into an ASR-like format and presented BERT-like models that solve the task of punctuation restoration efficiently, with our best model Hubert achieving a macro $F_1$-score of 82.2.


%
\renewcommand\bibname{References}
\bibliographystyle{splncsnat_en}
\bibliography{mszny}

\newpage
\appendix
\renewcommand{\thesection}{\Alph{section}.\arabic{section}}
\setcounter{section}{0}

\end{document}